\documentclass{article}

\usepackage[margin=1in]{geometry} 
\usepackage{lipsum}
\usepackage{enumitem}

\usepackage[utf8]{inputenc} 
\usepackage[T1]{fontenc}    
\usepackage{url}            
\usepackage{booktabs}       
\usepackage{amsfonts}       
\usepackage{nicefrac}       
\usepackage{microtype}      
\usepackage{graphicx}
\usepackage{natbib}
\usepackage{doi}
\usepackage{float}
\usepackage{amsmath}
\usepackage{amsthm}
\usepackage{multirow}
\usepackage{algorithm}
\usepackage{algpseudocode}
\usepackage{xcolor}
\usepackage{hyperref}
 
\hypersetup{
    colorlinks,
    linkcolor={red!50!black},
    citecolor={blue!50!black},
    urlcolor={blue!80!black}
}
\newcommand{\floor}[1]{\left\lfloor #1 \right\rfloor}
\setcitestyle{authoryear, open={(},close={)}}

\newcommand{\beginsupplement}{%
        \setcounter{table}{0}
        \renewcommand{\thetable}{S\arabic{table}}%
        \renewcommand{\theHtable}{S\arabic{figure}}
        \setcounter{figure}{0}
        \renewcommand{\thefigure}{S\arabic{figure}}%
        \renewcommand{\theHfigure}{S\arabic{figure}}
        \setcounter{section}{0}
        \renewcommand{\thesection}{S\arabic{section}}%
        \renewcommand{\theHsection}{S\arabic{section}}%
     }

\title{\bf{Language Model Memory and Memory Models for Language}}

\author{\small \href{https://orcid.org/0000-0003-1661-4579}{\color{black} \textbf{Benjamin L. Badger}} \thanks{The author would like to thank IBM for support during the writing of this paper. Code may be found on \url{https://github.com/blbadger/memorymodels}.} \\
	\small IBM \\
	\small \texttt{ben.badger@ibm.com} 
}

\date{}

\hypersetup{
pdftitle={Language Model Memory},
pdfsubject={Deep Learning},
pdfauthor={Benjamin L. Badger},
pdfkeywords={Deep Learning, Representation Theory, Memory},
}

\begin{document}
\maketitle

\begin{abstract}\normalsize{
    The ability of machine learning models to store input information in hidden layer vector embeddings, analogous to the concept of `memory', is widely employed but not well characterized. We find that language model embeddings typically contain relatively little input information regardless of data and compute scale during training. In contrast, embeddings from autoencoders trained for input regeneration are capable of nearly perfect memory formation. The substitution of memory embeddings for token sequences leads to substantial computational efficiencies, motivating the introduction of a parallelizable encoder-decoder memory model architecture. Upon causal training these models contain information-poor embeddings incapable of arbitrary information access, but by combining causal and information retention objective functions they learn to form and decode information-rich memories. Training can be further streamlined by freezing a high fidelity encoder followed by a curriculum training approach where decoders first learn to process memories and then learn to additionally predict next tokens.  We introduce the perspective that next token prediction training alone is poorly suited for accurate memory formation as the objective itself is non-invertible, motivating the use of combined objective functions for models where the entire input is not exposed.
    }
\end{abstract}

\section{Introduction}

    A fundamental feature of machine learning models is their ability to store and use information. The nature of this information storage and processing varies depending on the model in question, but for transformer-based large language models information is stored in the model's parameters as well as the activations of various layers. With some slight oversimplification, one can think of the information attained during training as existing in the model's weights and the information attained from any given input as existing in the model's activations. In this work we focus on the informational content language models posses from being exposed to an input token sequence, not its training data. For convenience we use the notion of the last hidden layer embedding of the last token of an input sequence and that model's `memory' of that sequence interchangeably.

    The substitution of memory embeddings for sequences of tokens gives clear benefits to many aspects of inference computation (such as lower time to first token and smaller KV cache sizes), and has been attempted in various forms because of this. To date memory models have shown promise for single-lookup (phonebook) style tasks \citep{bulatov2022recurrentmemorytransformer} and associative retrieval \citep{rodkin2025associativerecurrentmemorytransformer} and variable-length compression \citep{dai2019transformerxlattentivelanguagemodels}, but have not been widely adopted for general language tasks. In this work we provide a basis for why memory models have underperformed their full-context counterparts, namely that as they currently exist they are incapable of the arbitrary information access characteristic of full-context Transformers, which we argue is a necessary element in their emergent abilities for tasks they have not been explicitly trained for (see \citep{radford2019language}). Because of this we focus on memory model architectures designed with three primary considerations: next token prediction training efficiency, arbitrary input information storage and use, and parallelization. 
    
\subsection{Related Work}

    LLMs have been found to be invertible in certain scenarios: \citep{morris2023languagemodelinversion} observed accurate conversational prompt input regeneration using last hidden layers or even logits. We provide results to show that input sequence regeneration from output embeddings is much lower for larger context windows over more diverse language corpus, which may be predicted by observing the low out-of-distribution compared to in-distribution results in that work. 
    
    \citep{weller2025theoretical} found that single-embedding retrieval models far underperform both their theoretical potential as well as multi-embedding or classic search algorithms when matching a query to lists. From the perspective and results presented in this paper, we provide a basis for these observations: retrieval of one element in a list requires near-lossless compression of those list elements into the embeddings, but the amount of compression possible if each list element is randomly sampled is necessarily low.

    Numerous investigations into the use of embeddings as memories exist, most being motivated by the potential to extend context windows of models. One that is particularly relevant is \citep{bulatov2022recurrentmemorytransformer}, where the authors incorporate one or more embeddings from a causal decoder on a context sub-window into future sub-windows in a recurrent manner. Somewhat analogously, \citep{dai2019transformerxlattentivelanguagemodels} uses stop-gradient transformations to allow multiple hidden layer embeddings from past tokens to inform a model's prediction of future tokens. \citep{lin2025refragrethinkingragbased} optimizes retrieval-augmented generative search by augmenting causal decoder inputs with memory embeddings. Contemporaneously with this work, \citep{prakash2025attentioncompressionneedcontrollably} introduced a nearly identical parallelized memory model architecture to one of those presented in this work (the non-frozen transformer-based memory model) and explore the benefits of training this model at somewhat larger compute and data scales. \citep{sukhbaatar2021memoriescreatedequallearning} investigate the creation and destruction of memories for very long-context modeling with fixed memory usage.

    The primary point of departure in our work is the recognition that memory models should allow for near-arbitrary information lookup to be comparable with full-context transformers, which with the finding that causal training is generally unsuitable for high-fidelity memory formation necessitates a new training paradigm for these models.
    
\section{Our Contribution}

    This work seeks to address the following question:
    
    \textit{How accurate are causal language model memories, or in other words how much input information do these models retain?}

    \noindent and we answer this as "not very much". The need to access arbitrary input elements for modern language tasks makes such models unsuitable for providing memory embeddings to other models, which combined with the observation that autoencoders are capable of highly accurate memory formation motivates the following question:

    \textit{Can models be trained to form accurate memories, and can causal decoders learn to use information from perfect-memory encoders?}

    \noindent where we find that this answer is "yes". We in    troduce the following methods and approaches:
    
    \vspace{0.15cm}
    \begin{enumerate}[nosep]
    \item Information extraction methods via inversion for two architectures
    \item An information quantification approach for cross-model comparison
    \item Parallelizable memory model architectures with combined objective functions for accurate memory
    \item Frozen-memory architectures for arbitrary information retention via curriculum training
    \end{enumerate}

\section{Memory and Useful Memory}

    \subsection{Why Memory?}

    The most common application of deep learning language models today is autoregressive generation, where a sequence of tokens is fed to the model and tokens are predicted one at a time. In order to predict any single next token it may generally be assumed that only a small fraction of previous tokens are required, although it cannot be determined \textit{a priori} which tokens these will be due to the inherent freedom in language such that \textit{any} previous token's information may be requested to predict a next token. A prediction step for these models may be described as a function $f: X \to Y$ where $X \in \Bbb Z^n$ is the sequence of $n$ input tokens and $Y \in \Bbb Z^1$ is the single output token, where $f$ is by many-to-one function. If we disregard tokens and consider only embeddings of tokens $X \in \Bbb R^{n \times d}$ and output memory embeddings $Y \in \Bbb R^{d}$ we again have a many-to-one function, specifically one that maps sequences of vectors to vectors. Many-to-one functions are typically non-invertible and as such offer no guarantees of how much input information will be stored in any intermediate state. In our parlance, there is no guarantee that models trained to predict each next token will have good internal memory because they are exposed to the entire input and are optimized on an objective of predicting a single next token at a time, which normally requires the knowledge of only a handful of previous tokens. Because of this it can be argued that causal models are unlikely to form high-information embeddings, as they would not be expected to need to do so to minimize training loss.

    Understanding memory informational content is more important when models are not exposed to the entire input, if for example a model were to use an embedding in place of a sequence of tokens.  Memory model inference is made more difficult by the need to be able to extract any of the relevant information in memory embeddings, as this stipulation leads to the requirement that embeddings contain \textit{all} necessary input information because which tokens will and will not be used in the future is usually not predictable. The benefits of such models are greatly reduced computation at inference, explored in the next section.
    
    \subsection{Computational Complexity}
    
    The design of effective memory models is motivated by numerous computational efficiencies at inference: for a given sequence, memory models requires less cache memory and computation, leading to lower time-to-first-token, lower on-device memory, and higher token throughput. These models are expected to trade increased computation at training (as the extra step of storing and decoding memories must be learned as well as next token prediction) for this decreased inference computation. For Transformer memory models for $n$ tokens the computation for full-context models is proportional to $n^2$, whereas for a memory model embedding with $s$ sequences (`chunks') of $n/s$ tokens each the computation is proportional to $s(n/s)^2$ for the encoder and $s^2$ for the decoder. Upon solving for the minimum value of the maximum of these functions, we find that the ideal value of $s$ in terms of $n$ is $s = n^{2/3}$ which results in computation proportional to $n^{4/3}$. In this work we investigate memory models in which encoder forward passes may be parallelized during training and inference, which for full parallelization with the above assumptions reduces the encoder's complexity to be proportional to $n^{2/3} = s$ and does not change the decoder's complexity. In the case where global memory bandwidth is limiting (which is common for single sample inference) memory models offer another substantial speedup: rather than loading $n * d$ elements per layer per token predicted, $n^{2/3} * d$ elements are loaded for each token predicted assuming $s = n^{2/3}$. For this work we typically use $s < n^{2/3}$, specifically $s=4$ for $n=1024$ or $n=2048$, raising the computational complexity but lowering the number of cache elements loaded.
    
    \subsection{Theory: Model Invertibility and Information Retention}

    What does it mean for a model's memory of an input to be good or bad? We answer this question as follows: a good memory is one that is accurate, in that it can be used to accurately reconstruct that input. Equivalently, a good memory contains all the information necessary to uniquely identify any given input, and for information-rich inputs such as language the memories are necessarily also information-rich. Also equivalently, a good memory is one that allows the model to be inverted.

    Invertibility in the context of deep learning models is a concept subtle enough to require further explanation, which we explore here. In the usual notation, given a model $\theta$ and input $X$ we obtain an output embedding $Y$ as $Y = O(X, \theta)$. Strictly speaking, model $\theta$ is invertible if and only if that equality may be represented by a function $f: X \to Y$ that is bijective, meaning both injective (one-to-one) and surjective (onto). If there are some values of $Y$ to which $f$ does not map some input to the model is not onto, or if there are values of $Y$ that multiple values of $X$ map to then the model is not one-to-one; in either case it is not strictly invertible.

    It may be easily appreciated that typical language model architectures are not invertible in this strict sense because they are composed of functions that are themselves not invertible: for example, the Transformer architecture \citep{vaswaniattention} makes use of linear transformations in MLPs, normalization layers, and attention operations all of which are non-invertible (most easy to demonstrate is that they are non-injective). In this sense, none of the models or architectures presented in this work are invertible to the degree that when arbitrary inputs are generated to approximate language model outputs, there is generally no guarantee that the generated input will approximate the original in any way \citep{badger2025maskedmixerslanguagegeneration}.

    On the other hand, suppose $X$ is a non-infinite set such as a collection of training samples, and $Y$ the non-infinite set of model embeddings of these inputs: in this case, the model is invertible if and only if it is representable by a function $f: X_n \to Y_n$ that maps each input to a unique output, which we index by $n \in \{0, 1, ..., N\}$ for dataset size $N$. Testing for invertibility over a finite dataset requires only that the number of unique outputs equals the number of unique inputs, and over practically all dataset and model combinations used today this is almost guaranteed to be true due in part to the extremely high dimensionality of the outputs and the non-linear nature of $\theta$ (for measurements on image datasets, see \citep{badger2023adversarialexamplesdimensionalinvariance}). In this input-restricted sense, therefore, language models are all invertible.

    These considerations motivate a third definition of model invertibility: an invertible model is one in which for a given set of inputs, each model output may be mapped to the corresponding input in that set with a mapping that is computationally feasible to find and perform. If we allow the inverse function to be trainable and assuming the the inverse function mirrors the model in some way, it can be assumed that finding the inversion function will be much more difficult than actually performing it. Essentially we consider models invertible if an inversion function can be practically found. Equivalently, a model with a good memory is one where some method can be feasibly found that can accurately reconstruct any input on a given corpus given the memory embedding. Computational feasibility being in the eye of the beholder, we define it to be around 96 V100 hours (around 24 H100 hours) which is a maximum theoretical total of around $170 * 10^{18}$ FLOP. Training details are given in section \ref{methodsection}.

\section{Language Model Memory Measurement}

     Prior work \citep{badger2025maskedmixerslanguagegeneration} found that two foundational models exhibited different degrees of invertibility or equivalently information information retention using a gradient descent technique. Like most inversion methods that don't involve training, that method was found to be incapable of accurately inverting single embeddings. We replace the gradient descent with a trainable decoder model to provide a more powerful inversion method, inspired by \citep{morris2023languagemodelinversion}.

    \subsection{Information Measurement Methods}

    We first tested language model inversion for relatively small models over the same dataset that these models were trained on (reporting hold-out evaluation data), and to do so we use a subset of the relatively large and varied FineWeb dataset, FineWeb-edu \citep{penedo2024finewebdatasetsdecantingweb}, which is itself filtered from Common Crawl snapshots. The experimental approach we take is as follows: first a model is trained for a given language task, then its parameters are frozen and a trainable decoder is added, the decoder is trained to invert the model's embeddings over its training dataset, and a measurement of this inversion in terms of information retention is obtained. In order to apply this technique to models of different types, we use two metrics that yield tokenizer size-independent estimates: an entropy ratio that measures the fraction of information of the input the embedding contains, and a Hamming metric-based token identification accuracy metric.

    The rationale behind the use of the entropy ratio $H_r$ is as follows: cross-entropy loss is equivalent to the number of bits per token required to convert the models' distribution to the target, where the model's output is a sequence of vectors of size $\vert t \vert$, the size of the tokenizer. We now define upper and lower bounds for this metric: the upper bound is simply zero as a perfect model requires no change to its output distribution to match the target, and the lower bound we compute as shown in Equation \ref{eq1}, which is the average of the cross entropy between an informationless output (the uniform distribution $\mathcal{U}(|t|)$, which has maximum entropy) and sample $t$ from any target distribution. We then take the complement of the ratio of achieved loss to upper bound loss (Equation \ref{eq2}). The lower bound may also be thought of as normalizing for the differences between tokenizers of various sizes, as it is clearly more difficult to guess a token from a very large tokenizer than one with only a few possible tokens which is reflected in the denominator of (\ref{eq2}). For the 8k-size tokenizer most commonly used in this work, (\ref{eq2}) simplifies to $H_r  = 1 - H(p, q)/9.03$. 

    \begin{equation}
        H(p_0, q) = \frac{1}{n} \sum_{n} \Bbb L \left( \mathcal{U}(|t|), t \right)
        \label{eq1}
    \end{equation}
    
    \begin{equation}
    \begin{split}
        H_r &= 1 - \frac{H(p, q)}{H(p_0, q)} = 1 - \frac{- \sum_x q(x) \log (p(x))}{- \sum_x q_0(x) \log (p(x))} \\
        \label{eq2}
        \end{split}
    \end{equation}

    We also use a Hamming metric-based measurement that provides the proportion of input tokens that the decoder has successfully inverted, which in words is the complement of the fraction of non-pad tokens that are correctly identified, and formally given in Equation \ref{eq3} which we typically report as token accuracy in terms of \% correct for disambiguation. This metric can be viewed as a somewhat higher bar than $H_r$, as containing a certain number of bits per token may or may not lead to its accurate identification.

    \begin{equation}
        h(x, y) = 1 - \frac{1}{j} \mathrm{Card} \left( \{ x_i \neq y_i \} \right) : x_i \neq t_{pad}, i \in 0, 1, ... , j
        \label{eq3}
    \end{equation}

    \subsection{Information Retention} \label{infosection}

    We test the amount of information contained in a model as follows: given an arbitrary model $M$, we first convert this model to a form appropriate for obtaining an embedding (removing language modeling or classification heads as necessary), such that it becomes an encoder mapping input tokens to a single embedding, $\mathcal E: \Bbb Z^{n} \to \Bbb R^d$. We then freeze the parameters of $\mathcal E$ and add a decoder (usually of the same size and architecture as the encoder) $\mathcal D : \Bbb R^{n \times d} \to \Bbb Z^n$ that takes as input he encoder's last hidden layer, last token embedding $\mathcal {E}(X, \theta)_{[:, -1]}$. The encoder's embedding is unrolled via a trainable projection which maps $P: \Bbb R^d \to \Bbb R^{n \times d}$ after \citep{morris2023languagemodelinversion}. The unrolling process is described in more detail in \citep{badger2025knowlimitsentropyestimation} and essentially amounts to applying a trainable linear projection from a sliding (in the embedding dimension) window of the the encoder's embedding to form all decoder inputs, and results in more efficient training than repeat or single-embedding introduction to the decoder. During our control experiments we found that replacing the encoder's frozen token embedding layer with a trainable transformation results in less decrease to validation model accuracy, and this is performed unless otherwise noted. The complete encoder-decoder information retention model is described by Equation \ref{eq4} and is depicted in Figure \ref{fig1}. 
    
    \begin{equation}
        Y = \mathcal {D}( P(\mathcal {E}(X, \theta)_{[:, -1]}), \theta_{\mathcal D})
        \label{eq4}
    \end{equation}
    
    \subsection{Causal Language and Retrieval Models but not Autoencoders exhibit Poor Memory}

\begin{figure}[h]
        \centering
        \includegraphics[width=0.99\textwidth]{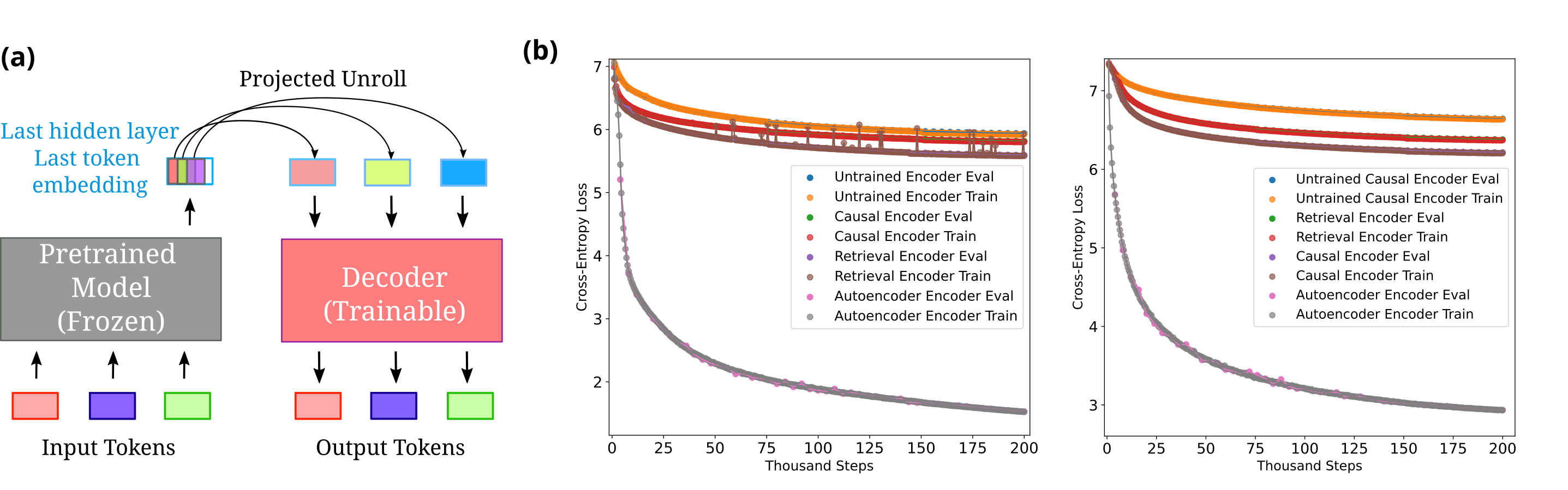}
        \caption{Information retention experimental approach (left) and example training runs (right, Masked Mixer on the left and Transformer on the right).}
        \label{fig1}
    \end{figure}

    Upon training encoder-decoder information retention models, we find that encoder loss tends to plateau and decrease only marginally with more training samples as shown in Figure \ref{fig1}. We find that models trained for next token prediction (causal modeling) and retrieval exhibit little increase in information retention compared to untrained models, but that autoencoders trained for input regeneration have substantially more information (Table \ref{table1}). We confirmed that the decoder-only training approach for information retention measurement does not lead to substantial under-estimation of information retention in pretrained autoencoders, where the encoder's information content is known (Figure \ref{figs1}, Table \ref{table1}).

    \begin{table}
    \renewcommand{\arraystretch}{1.3}
    \centering
    \begin{tabular}{|l|l|l|l|l|l|l|}
    \hline
        & \multicolumn{3}{|c|}{Masked Mixer} & \multicolumn{3}{|c|}{Transformer} \\
        Encoder Model & Loss &$H_r$ & Accuracy  & Loss & $H_r$ & Accuracy \\ \hline
        \hline
        Autoencoder ($n_{ctx}=256$ validation) & 0.01 & 0.999 & 99.6 \% & 0.09 & 0.99  & 95.4 \% \\ \hline
        Autoencoder (validation) & 0.292 & 0.968 & 91.84 \% & 2.73 & 0.698 & 43.3 \%  \\ \hline
        Autoencoder  & 0.435 & 0.952 & 76.96 \% & 2.935 & 0.624 & 33.5 \% \\ \hline
        Untrained & 5.937 & 0.343 & 4.67 \% & 6.643 & 0.264 & 4.67 \% \\ \hline
        Causal & 5.815 & 0.356 & 5.48 \% & 6.214 & 0.31 & 5.01 \% \\ \hline
        Causal -> Retrieval & 5.594 & 0.381 & 5.45 \% & 6.380 & 0.29 & 5.29 \%  \\ \hline
        Autoencoder -> Retrieval & 5.64 & 0.375 & 5.46 \% & - & - & - \\ \hline
        \end{tabular}
    \caption{Information Retention (all models $n_{ctx}=512, n_l=8$ mixers $d_m=1024$ and transformers $d_m=512$, trained for 200k steps). `Causal -> Retrieval' signifies a model first pretrained for causal modeling, then trained for retrieval via InfoNCE. Validation models were trained as autoencoders once, rather than trained followed by a freeze on the encoders and re-trained.}
    \label{table1}
    \end{table}

    \subsection{Autoencoder Memories are Non-Trivial}

    We reasoned that the high memory fidelity, or equivalently high compression of autoencoders relative to causal and retrieval models could be caused by the encoder learning to embed the input in a non-data-dependent manner, which we call a `trivial' autoencoding. One example of such an encoding would be possible for any model where the encoder's embedding dimension is at least as large as the number of tokens in the input, $d_m\geq n_{ctx}$, in which case the encoder could learn to simply encode one token per embedding element. We tested the data dependence of trained autoencoders by observing the cross-entropy loss of these models applied to sequences of uniform random tokens, and observe that the this loss is far higher than the in-distribution loss (Figure \ref{figs3}). We further observe that trivial embeddings are not trainable with our autoencoders and optimization methods: training autoencoders on sequences of uniform random tokens results in no optimization (Figure \ref{figs3}). We also find that autoencoders trained on FineWeb generalize well to marginally out-of-distribution FineMath inputs (Figure \ref{fig3}).
    
    \subsection{Larger LLM Memory}

    We tested whether large increases in model size and increased pretraining compute and data would result in greater information retention by training decoders using embeddings from off-the-shelf LLMs. We chose to investigate BERT large \citep{devlin2019bertpretrainingdeepbidirectional} as this model contains around five times the number of parameters as the transformers presented in the last section and was trained with 10x the compute, as well as Qwen 3 0.6b which is 9x the size of the transformers we trained \citep{yang2025qwen3technicalreport} and Llama3.1 1b at 14x the size \citep{grattafiori2024llama3herdmodels}, both trained with >1000x the compute. For this experiment, we freeze the pretrained model's token embedding layer and model layers (without a language modeling head) and feed the last hidden layer output to a decoder after unrolling. As shown in Table \ref{table3}, there is a modest increase in token accuracy and entropy ratio for the same context as observed previously ($n_{ctx}=512$).
    
    These results contrast with those obtained in \citep{morris2023languagemodelinversion}, where conversational prompts were able to be recovered from causal LLMs. We hypothesized that there may be two reasons for this; firstly that shorter ($n_{ctx} \leq 64$) inputs should be easier to invert as they contain less information, and secondly that less-diverse inputs should be easier to invert. We tested the first hypothesis by observing input regeneration for smaller contexts, and find an inverse correlation between prompt length and regeneration accuracy (Table \ref{table3}). The second hypothesis was tested by training decoders to invert BERT outputs on a math-specific subset of the FineWeb, FineMath 4+ \citep{allal2025smollm2smolgoesbig}, and we observe nearly double the token recovery accuracy per context window on this dataset.

    \begin{table}
    \renewcommand{\arraystretch}{1.3}
    \centering
    \begin{tabular}{|l|l|l|l|l|}
    \hline
        Model & $n_{ctx}=16$ & $n_{ctx}=64$ & $n_{ctx}=256$ & $n_{ctx}=512$ \\ 
        \hline \hline
        BERT large & 0.719 (40.2\%) & 0.481 (11.6 \%) & 0.406 (6.6 \%) & 0.376 (5.96 \%) \\ \hline
        Llama 3.1 1b & 0.665 (34.8 \%) & 0.490 (11.2 \%) & 0.434 (6.0 \%) & 0.419 (5.30 \%) \\ \hline
        Qwen 3 0.6b & 0.614 (35.7 \%) & 0.459 (12.7 \%) & 0.414 (6.62 \%) & 0.402 (5.61 \%) \\ \hline
        BERT large (FineMath) & 0.891 (72.3 \%) & 0.641 (26.8 \%) & 0.528 (12.6 \%) & 0.491 (9.6 \%) \\ \hline
    \end{tabular}
    \caption{FineWeb (unless otherwise noted) LLM information retention for the given context windows. Entropy ratio $H_r$ and token accuracies in terms of \% correctly identified are reported.}
    \label{table3}
    \end{table}

\section{Memory Models}
    
    \subsection{Architecture and Implementation}

    Seeking to parallelize where possible, we introduce a memory architecture whereby an encoder is responsible for mapping sequences of tokens to an embedding, and a decoder takes any number of these embeddings in addition to any additional tokens of a sequence. These models have the notable advantage over decoder-only models such as \citep{bulatov2022recurrentmemorytransformer} in that they are capable of parallelizable processing of input sequences at inference, as the formation of any single embedding requires no input from any other embedding such that the encoder may be applied to all $s$ chunks in parallel. We train these models for causal modeling and to control for differences in decoder context length, we also train memory models without embeddings (embedding are replaces with tensors of ones).

     \begin{figure}[h]
        \centering
        \includegraphics[width=0.85\textwidth]{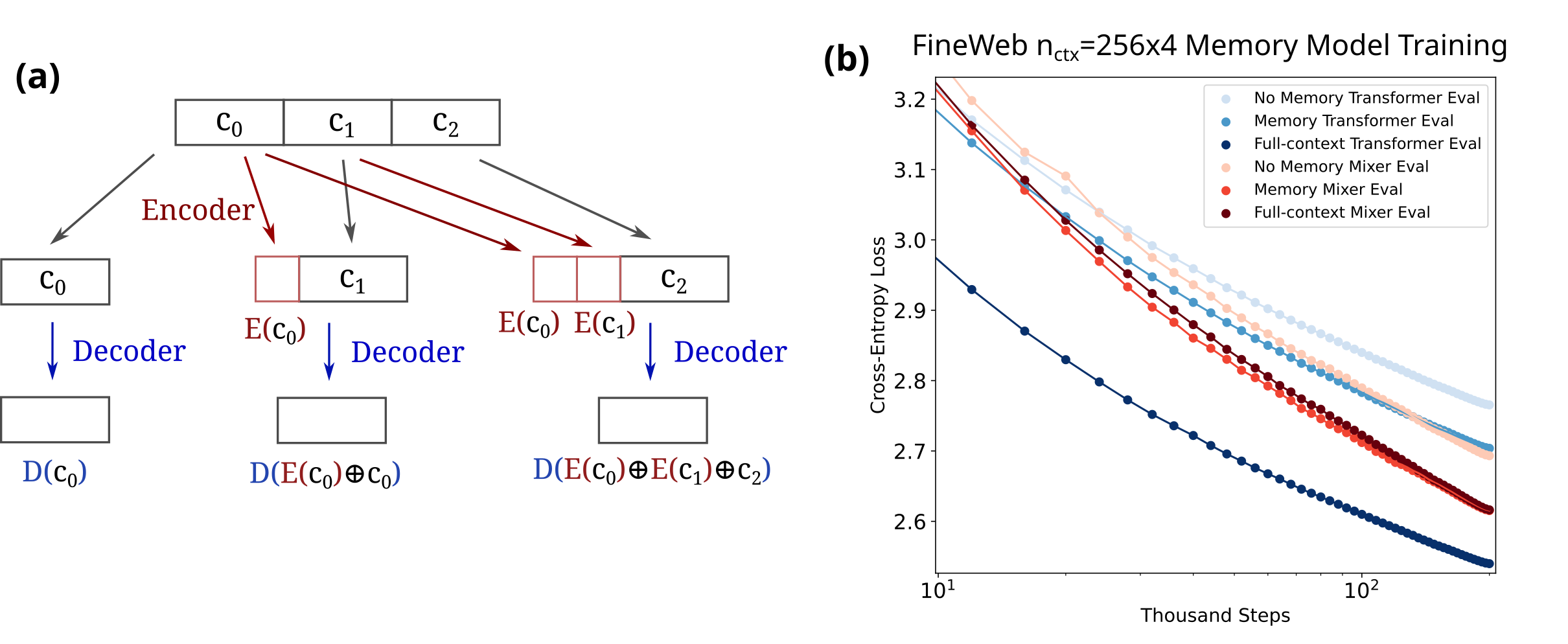}
        \caption{Memory Model Architecture and $n_{ctx}=256$ per chunk, $s=4$ chunk causal training characteristics on FineWeb. Mixers are $d_m=512$ for encoders, $d_m=1024$ for decoders and Transformers $d_m=256$ and $d_m=512$ for compute equivalence.}
        \label{fig4}
    \end{figure}

    We explore two varieties of memory model embedding introduction, one where a variable number of memory embeddings are introduced and token embeddings follow immediately, and the other where memory embeddings are introduced and padded such that the embeddings from the same memory chunks are always found in the same decoder positions. We find little difference in causal training efficiency, and in the rest of this work use the fixed embedding approach (Figure \ref{figs4}).

    \subsection{Causal Language Training Efficiencies of Memory Models}

    Causal training efficiencies of memory models are given in Figure \ref{fig4}, and we find that memory models can be nearly as efficient to train as full-context models but that this is somewhat architecture- and context window chunk size- dependent. We compared the causal training efficiencies of this parallelized memory model to that from an architecture where decoders recurrently form and process memory embeddings introduced by \citep{bulatov2022recurrentmemorytransformer}, which we implemented for both mixers and transformers with a single memory embedding per segment. Hypothesizing that a division between information retention and information extraction at the architectural level would assist training efficiency, we find that the parallelized encoder-decoder memory models are slightly more efficient to train on FineWeb than decoder-only recurrent memory models (Figure \ref{figs6}). 
    
    \subsection{Frozen Encoder Memory Models Train Efficiently Too}

    If the task of a memory model encoder is to capture as much information about their inputs as possible, we can separate encoders from decoders by first training the encoder to capture information and then training the decoder to predict next words using that information. This allows for increased parallelism and throughput during training, as frozen encoders don't form gradients and can be applied iteratively. As shown in Figure \ref{fig5}, with frozen autoencoder encoders memory model training efficiency approaches that of memory models with both encoder and decoder trainable. 

    \subsection{Pretrained LLMs as Memory Model Decoders}

    We next investigated whether or not it would be possible to adapt an LLM trained for next token prediction using token embeddings to use memory model embeddings as well. Using Llama 3.2 (1b) \citep{grattafiori2024llama3herdmodels} as the memory model decoder, we trained autoencoders to use the Llama 3 tokenizer, attached encoders from these models to trainable decoders and trained on the $n_{ctx}=256, s=4$ copy task depicted in Figure \ref{fig7}. We find that these models require the use of relatively small learning rates to prevent divergence, apparently due to difficulties backpropagating gradients through the entirety of Llama 3.2, and train relatively slowly (except for an initial drop in loss if trainable encoders are used) but do minimize the copy loss metric as shown in Figure \ref{figs7}. The use of a frozen information-rich encoder results in worse initial copy accuracy but faster convergence. We next investigated whether memory model training would interfere with the abilities Llama 3.2 gained during pretraining via benchmark evaluation of the decoder: we find that benchmark accuracy decreases somewhat after memory model training for most benchmarks tested but is most affected for pretrained (non-input) information retention as measured by MMLU accuracy (Tables \ref{tables4} and \ref{tables5}).
    
\section{Combined objectives for Accurate Memories}
    
    To be able to make use of arbitrary input information, memory models that use embeddings in place of token sequences must be capable of both encoding this information into embeddings and processing it in the decoder, and we test the potential for memory models to learn to do so in this section.

    \subsection{Experimental Setup and Evaluation}

\begin{figure}[h]
        \centering
        \includegraphics[width=0.99\textwidth]{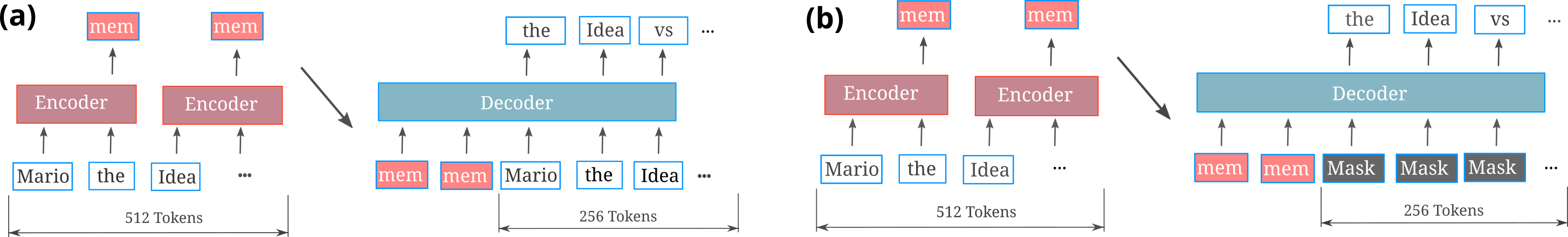}
        \caption{(a) Copy and (b) Blank Copy experimental schematics}
        \label{fig7}
    \end{figure}

    We use three methods for evaluating memory accuracy in memory models: the encoder-decoder information retention test detailed in Section \ref{infosection}, a copy objective originally introduced by \citep{hochreiter1997lstm} in which a sequence of tokens is sampled from the dataset, the first half is copied to the second, losses are masked on the first and the model is evaluated on its ability to accurately predict the copied tokens (having seen the correct tokens previously), and what we term a `blank copy' objective in which the tokens are masked, and only memory embeddings are available for the decoder when attempting to predict next tokens (Figure \ref{fig7}).  The copy task as applied to a memory model is a test of the model's ability to transfer sufficient information from the first $s/2$ chunks for a $s$ chunk model to the decoder to predict each single next token, given than the first half of the sequence contains all information necessary to perform this prediction. This is notably a less information-intensive task than the autoencoding information retrieval task because all previous tokens not encoded by a memory embedding are given to the decoder in its attempt to predict token $n$, such that a model with no information retention ability achieves a copy accuracy equivalent to its next token prediction accuracy (which from Table \ref{table4} is around 42\%) whereas an autoencoder with no information retention ability achieves near-zero input reconstruction accuracy. The blank copy task is designed to require the same informational content as encoder-decoder information. In our implementations, we copy sequences of 512 input tokens on-demand in the forward pass.

    The difference between the information measured by blank copy (or encoder-decoder retention models) and copy tasks may be summarized as follows: blank copy information is equivalent to the ability to the average per-token information necessary to construct the entire input sequence from scratch, and the copy metric is the average per-token information necessary to add a single arbitrary next token to the input sequence given all previous tokens.

    \subsection{Causal Memory Model Embeddings have Little Information}

\begin{figure}[h]
        \centering
        \includegraphics[width=0.99\textwidth]{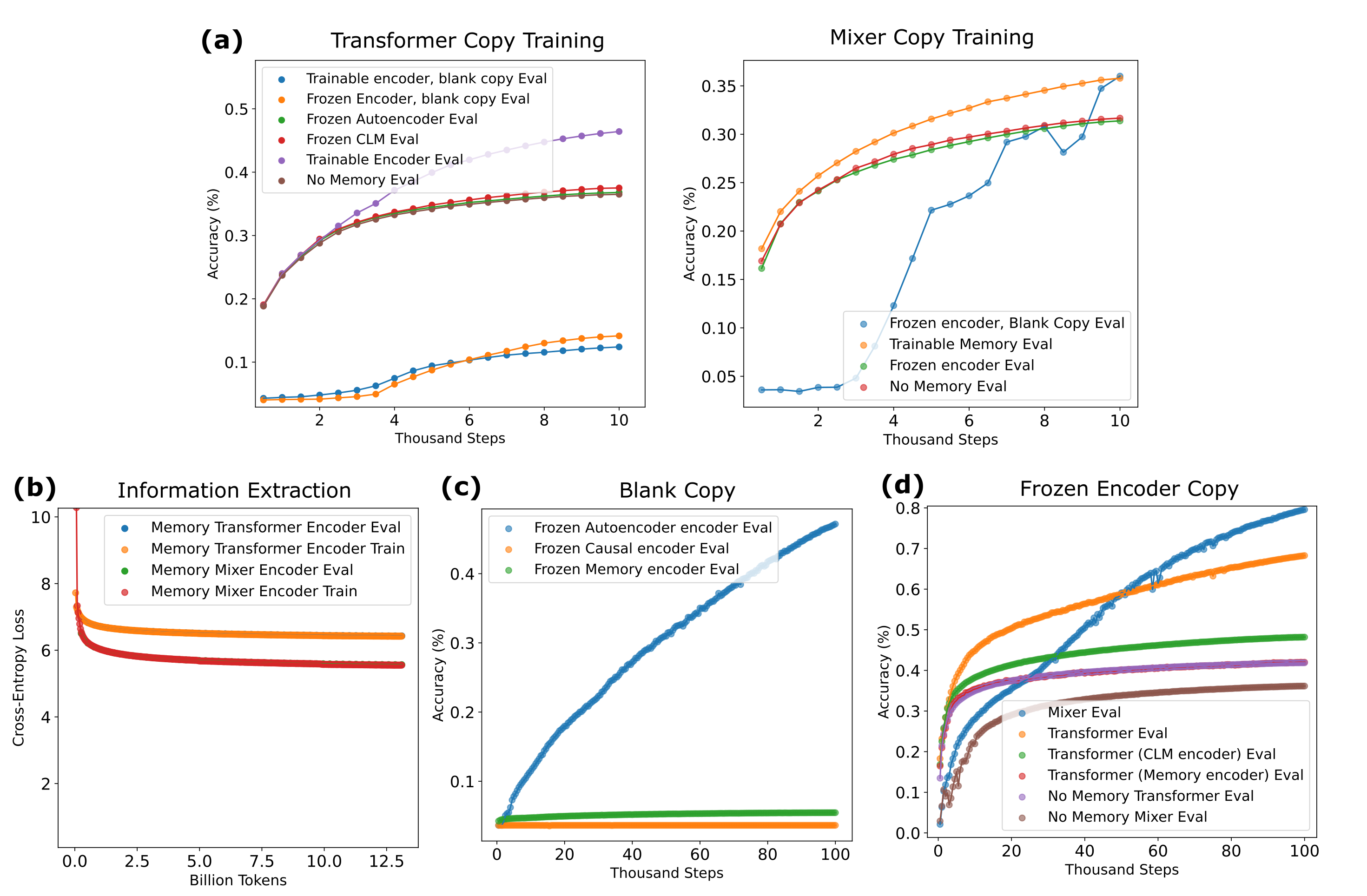}
        \caption{Memory Model Information Content and Copy Training. (a) Memory model training on the copy task. (b) $d_m=256, n_{ctx}=512$ Transformer and $d_m=512, n_{ctx}=256$ Mixer memory model encoder information, (c) Memory Transformer training on the blank copy task, (d) Frozen encoder copy training, all curriculum trained except CLM encoder and Memory encoder. All models $d_m=512, n_l=16, n_{ctx}=256, s=4$ unless otherwise noted and copy 512 tokens.}
        \label{fig8}
    \end{figure}

\begin{figure}[h]
        \centering
        \includegraphics[width=0.99\textwidth]{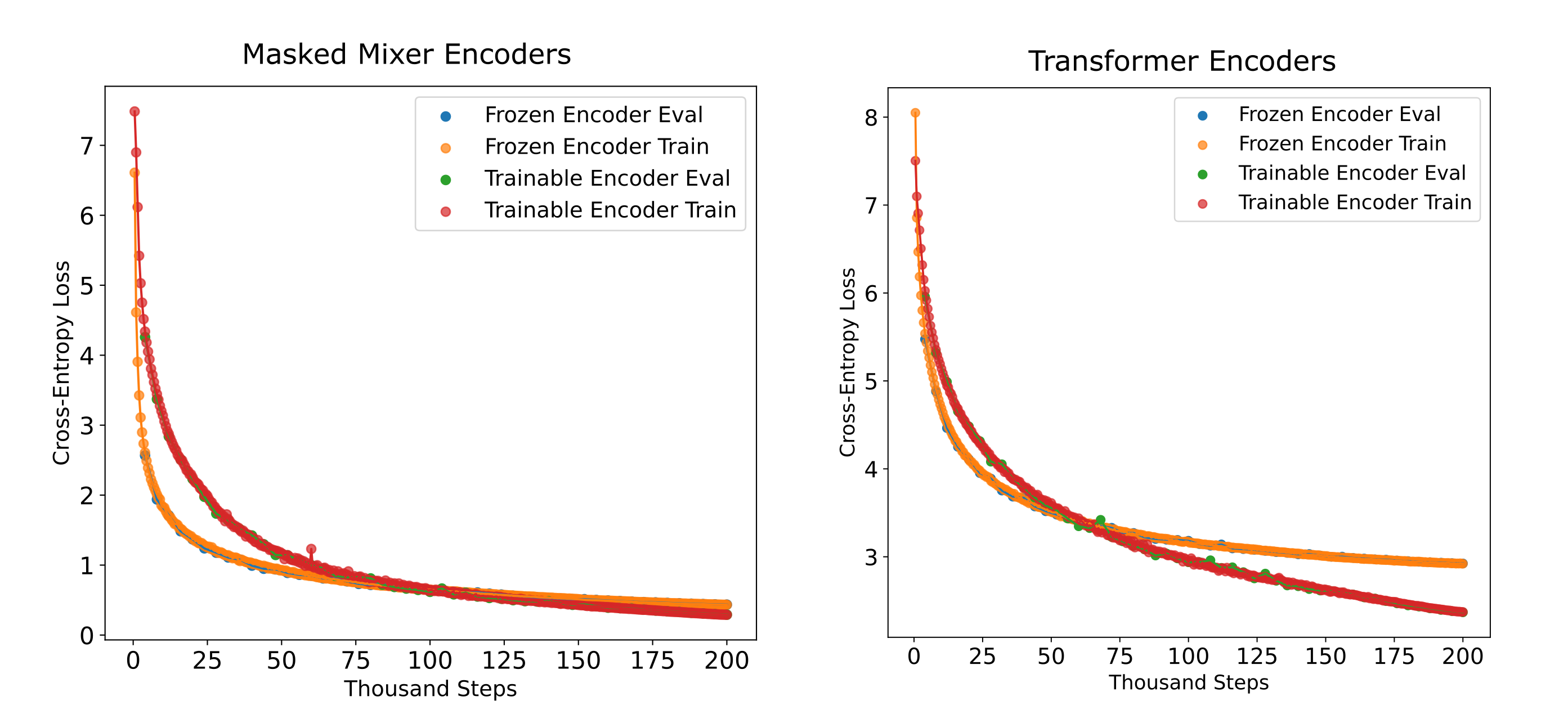}
        \caption{Mixer and Transformer information retention validation training curve examples on FineWeb. Transformers are $d_m=512, h=4$ and mixers $d_m=1024$ non-headed, both with $n_{ctx}=512, n_l=16$.}
        \label{figs1}
    \end{figure}

\begin{figure}[h]
        \centering
        \includegraphics[width=0.99\textwidth]{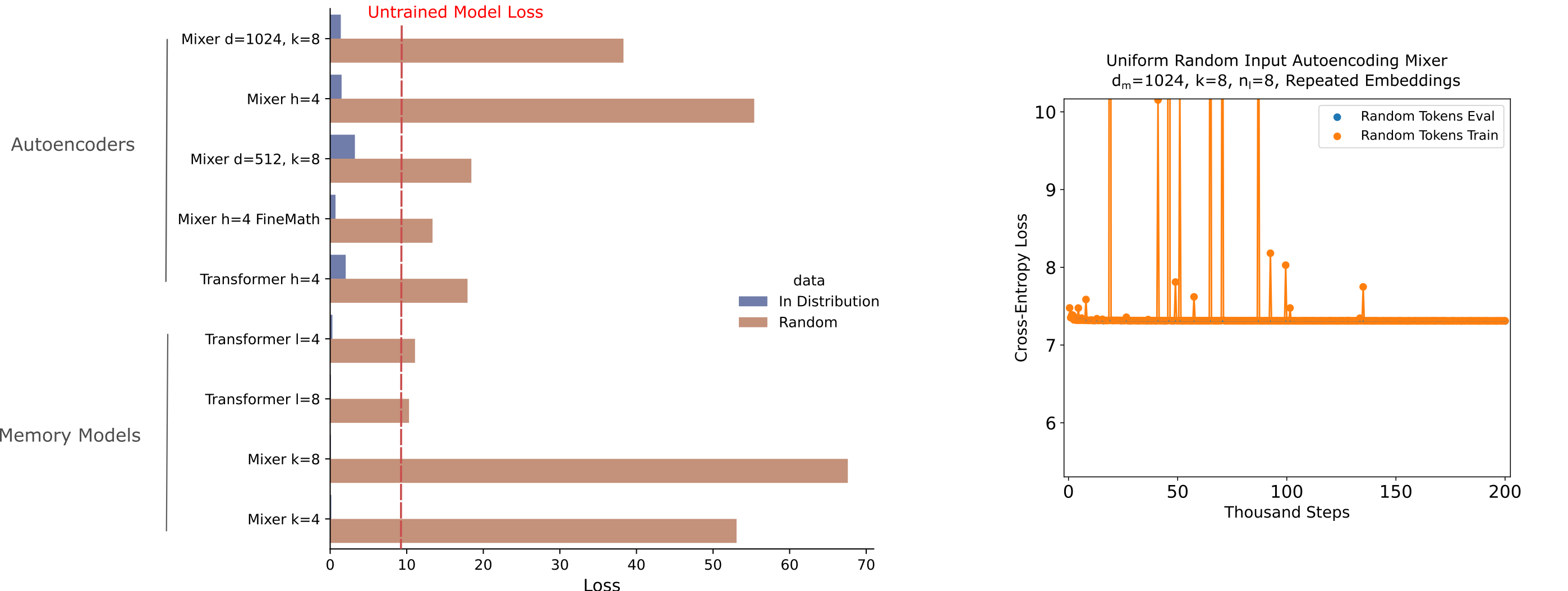}
        \caption{Autoencoders and Memory Model Encoder distribution dependence. Left, Autoencoder and Memory Model generalization to OOD uniform random distribution inputs versus hold-out FineWeb in-distribution inputs. Right, Masked Mixer autoencoders do not learn to model uniform random inputs. Memory models are `oracle' memory models where a full-context encoding is added to a causal decoder and trained for next token prediction, resulting in very small loss.}
        \label{figs3}
    \end{figure}

\begin{figure}[h]
        \centering
        \includegraphics[width=0.99\textwidth]{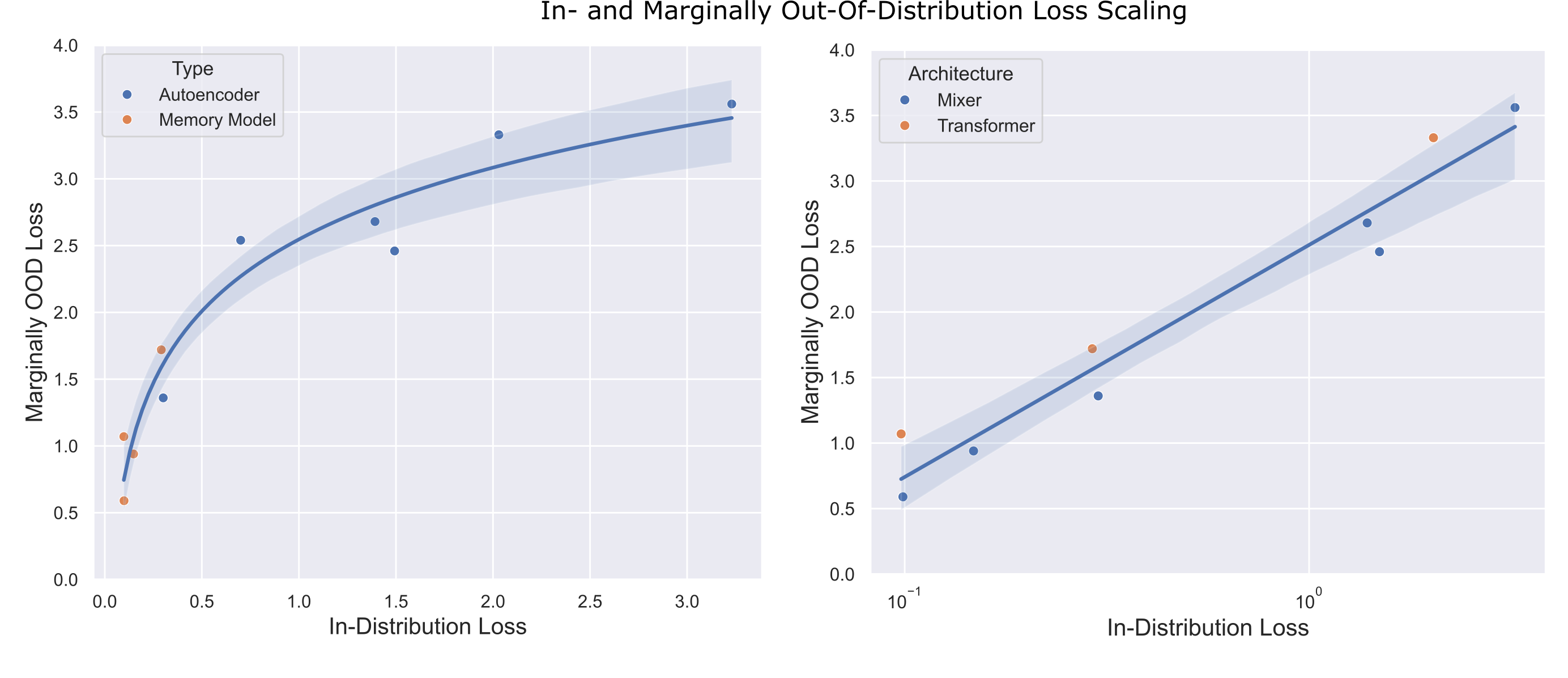}
        \caption{Autoencoder and Memory Model (oracle) encoder generalization. Models were trained on FineWeb-edu with FineMath 4+ used for marginally out-of-distribution samples. }
        \label{fig3}
    \end{figure}

\begin{figure}[h]
        \centering
        \includegraphics[width=0.99\textwidth]{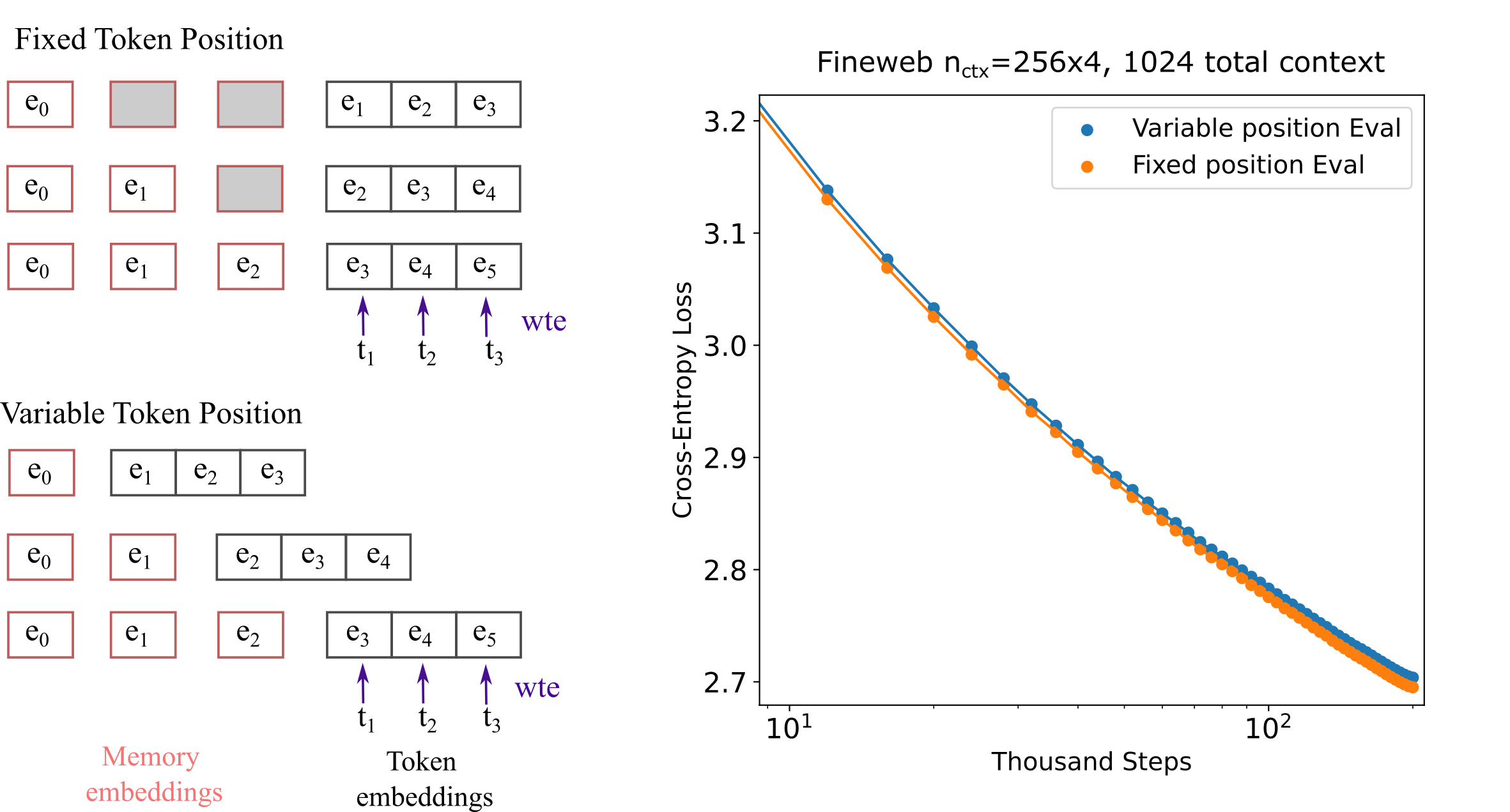}
        \caption{Fixed versus variable position memory Transformer training on FineWeb.}
        \label{figs4}
    \end{figure}

\begin{figure}[h]
        \centering
        \includegraphics[width=0.99\textwidth]{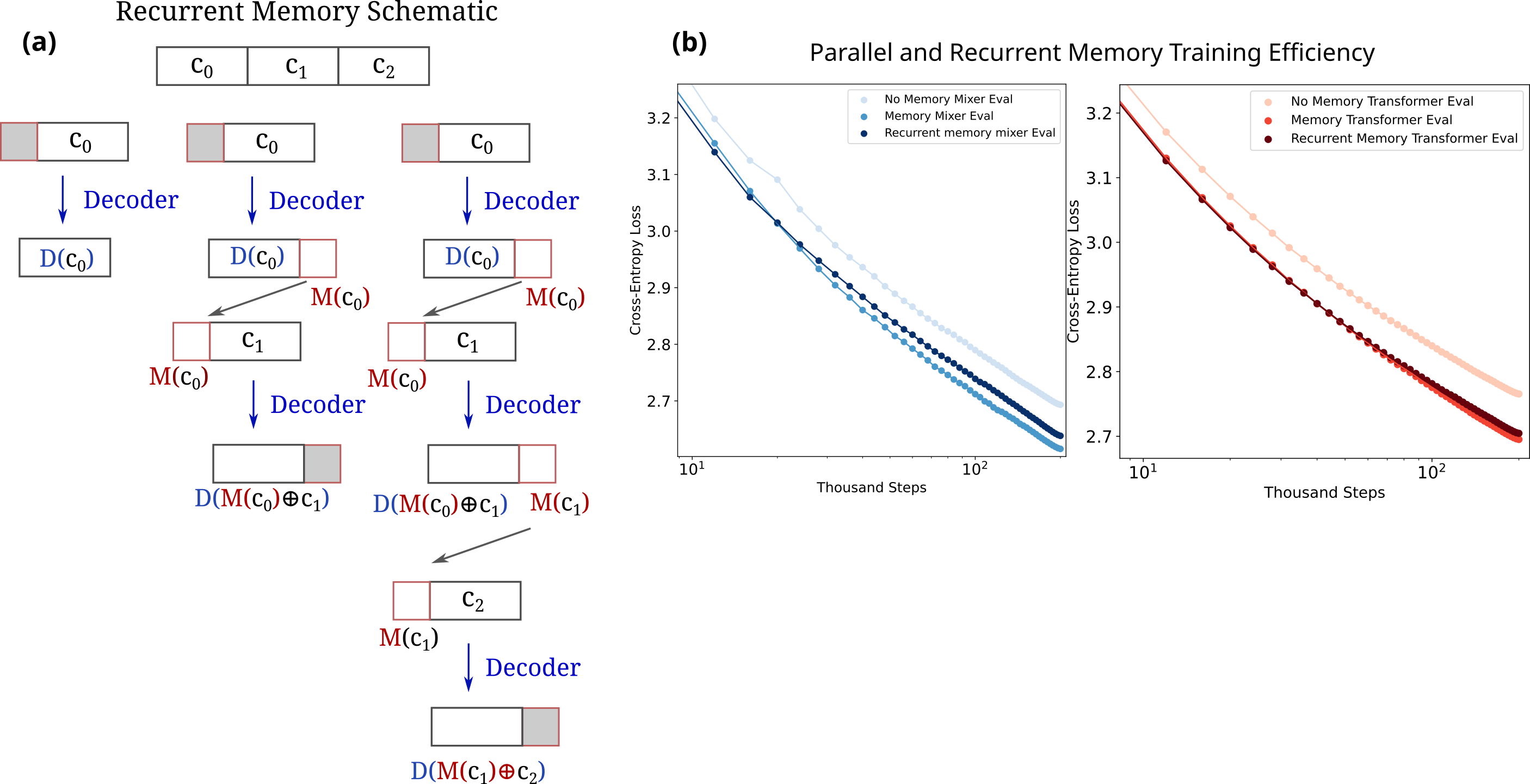}
        \caption{Parallel and recurrent memory model training efficiency on FineWeb.}
        \label{figs6}
    \end{figure}

\begin{figure}[h]
        \centering
        \includegraphics[width=0.8\textwidth]{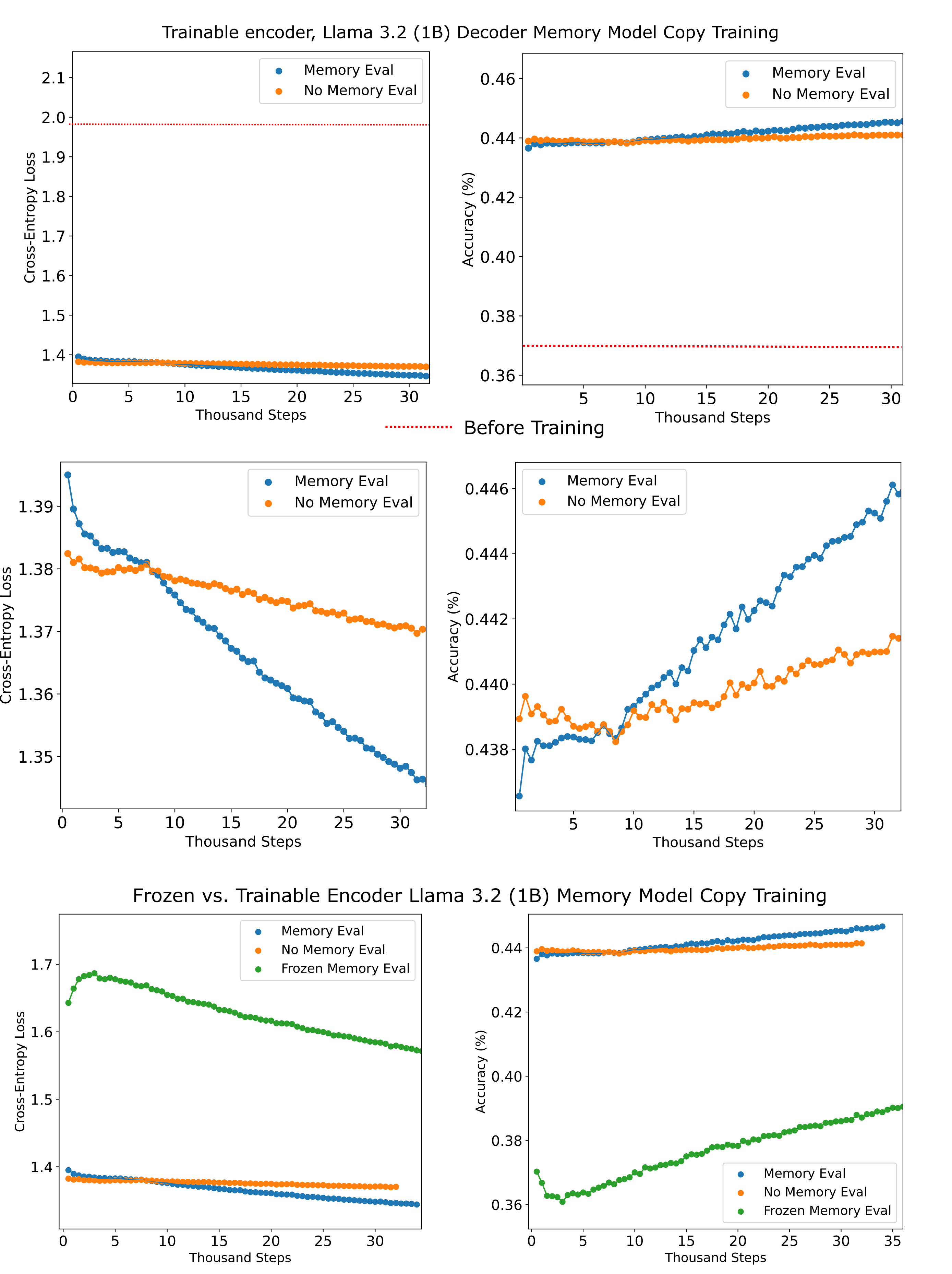}
        \caption{Llama 3.1 (1b) decoder memory model copy training. Models are $n_{ctx}=256, s=4$ with 512 tokens copied from FineWeb sampled inputs.}
        \label{figs7}
    \end{figure}

\begin{figure}[h]
        \centering
        \includegraphics[width=0.99\textwidth]{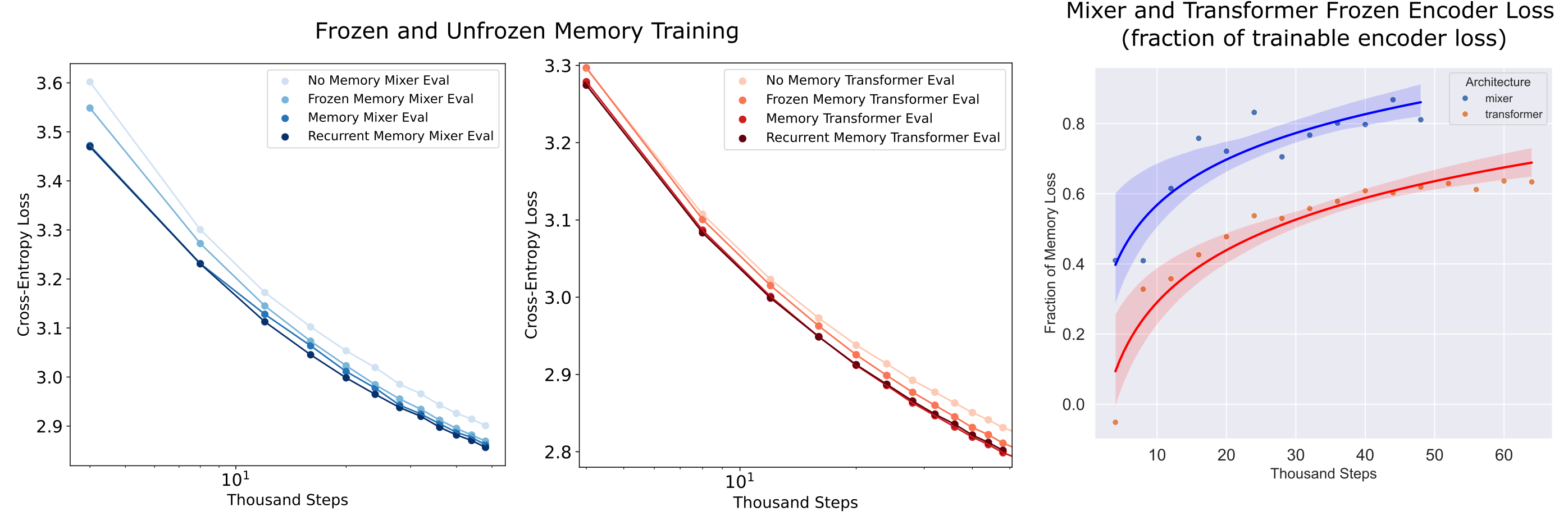}
        \caption{Memory model ($n_{ctx}=512$, $s=4$ chunks) causal training on FineWeb.}
        \label{fig5}
    \end{figure}

    As we observed relatively little information in causal decoder embeddings, we investigated whether encoders from causal-trained memory models would too. We first measured this via copy task training in which the memory model encoder is frozen and the decoder trained, before measuring performance on the more information-intensive encoder-decoder information extraction and blank copy tasks. We find that causal-trained memory models exhibit minimal increases in copy accuracy over the non-memory controls (Figure \ref{fig8} (d)), do not exhibit substantially higher information retention than causal full-context decoders (Figure \ref{fig8} (b)), nor are they capable of learning the blank copy task (Figure \ref{fig8} (c)). 
    
    \subsection{Curriculum Learning for Decoding Frozen Encoder Memory}

\begin{figure}[h]
        \centering
        \includegraphics[width=0.79\textwidth]{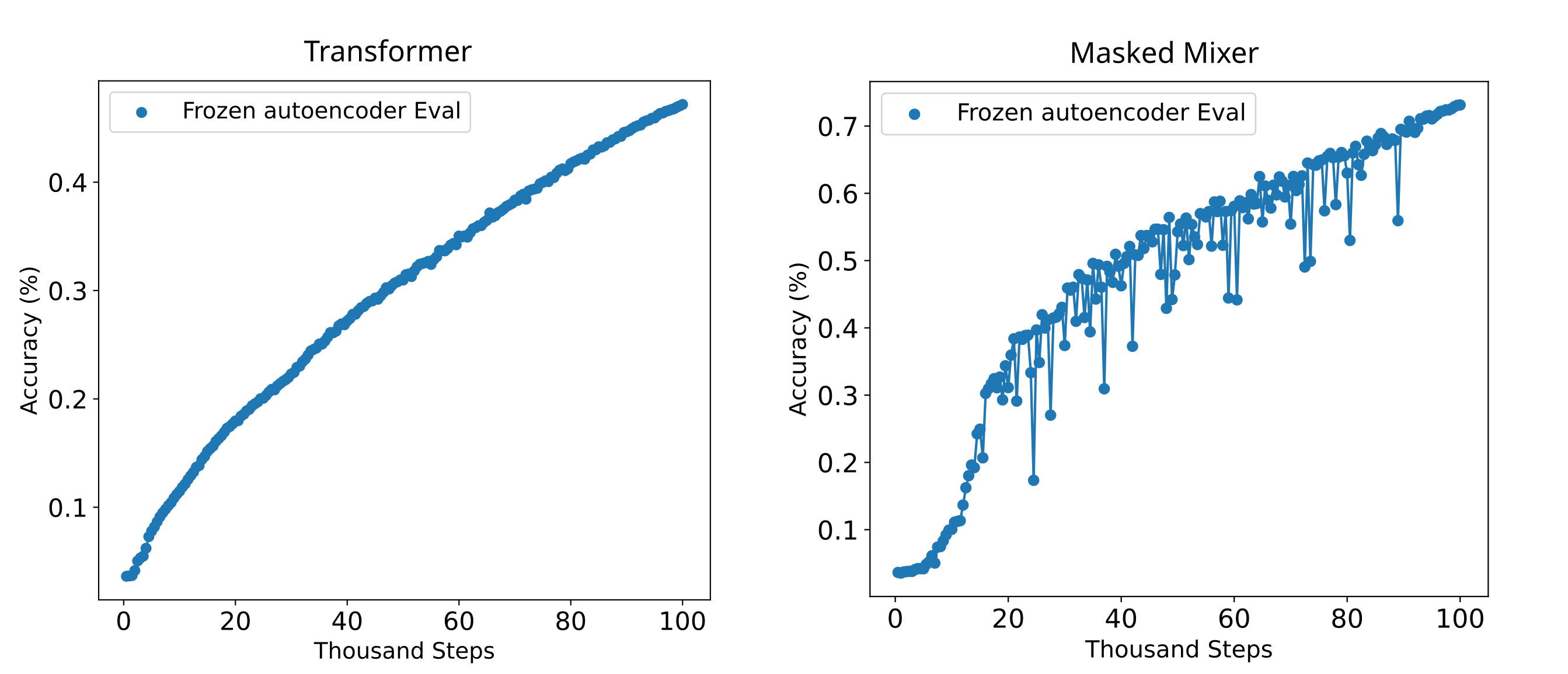}
        \caption{Blank Copy Training on FineWeb, 512 copy tokens, $d_m=512, n_l=16, n_{ctx}=1024$}
        \label{figs10}
    \end{figure}

    We were surprised to observe that frozen autoencoder encoder embeddings conferred no benefit to copy model training (Figure \ref{fig8} (a)), despite these embeddings containing a near totality of input information (Table \ref{table1}). Reasoning that this might be because decoding tokens (specifically predicting next tokens while disregarding memory embeddings) may be easier than decoding information-rich embeddings leading to models reaching local minima during training, we tested a curriculum learning approach where memory model decoders first learn to process information-rich memories without token embeddings followed by introduction of token embeddings. After training frozen autoencoder encoder memory models on the blank copy task (Figures \ref{fig8} (c), \ref{figs10}) we then trained the models on the copy task and find that copy accuracy further increases (Figure \ref{fig8} (d)), indicating that the memory model decoders are capable of using sufficient information from frozen autoencoder encodings for copying. In contrast, causal-trained encoders are unsuitable for this curriculum approach as they are poorly trainable to copy without token information (Figure \ref{fig8} (c)) and result in a modest increase in copy accuracy when trained without this curriculum (Figure \ref{fig8} (d)). 

    \subsection{Causal and Copy Combined Objectives for Accurate Memory Formation}

    \begin{table}
    \renewcommand{\arraystretch}{1.1}
    \centering
    \begin{tabular}{|l |l l|l l| l l | l |}
    \hline
         Objective & \multicolumn{2}{|c|}{Combined} &  \multicolumn{2}{|c|}{CLM} & \multicolumn{2}{|c|}{Combined (Blank)} & \\ 
         Metric & $H_r$ & Accuracy & $H_r$ & Accuracy & $H_r$ & Accuracy & \\ 
         \hline \hline
        CLM Eval & 0.675 & 42.07 \% & 0.684 & 43.15 \% & 0.669 & 41.5 \% & \multirow{2}{*}{All Trainable}\\ 
        Copy Eval & 0.947 & 86.1 \% & 0.688 & 43.65 \% & 0.756 & 50.6 \% & \\ \hline 
        CLM Eval & 0.664 & 40.8 \% & 0.677 & 42.32 \% & 0.660 & 40.4 \%  &\multirow{2}{*}{Frozen Encoder}\\ 
        Copy Eval & 0.927 & 82.7 \% & 0.677 & 42.63 \% & 0.759 & 52.6 \% & \\ \hline
    \end{tabular}
    \caption{Combined and causal trained memory model performance on causal and copy tasks.}
    \label{table4}
    \end{table}

    Given that memory models are capable of next token prediction and accurate memory formation in isolation, we investigated whether these models are capable of both conjunction. As memory models trained for next token prediction alone do not tend to form or make use of accurate memories, we train memory models using a linear combination of cross-entropy loss of next token prediction and copy training ($X_s$ denotes shifted inputs, $X_c$ denotes copied inputs with a copy delimiter added for disambiguation, and $X_{sc}$ shifted copied inputs) given in Equation \ref{eq10}. As shown in Table \ref{table4}, memory models are capable of learning to accurately copy and predict next tokens. For the more difficult task of copying without token provided, training via the combined objective function yields lower copy and causal accuracies, and in this case the curriculum trained frozen encoder results in higher accuracy than memory models with trainable encoders (Table \ref{table4}). Training memory models to form and decode information-rich embeddings results in a modest decrease in causal training efficiency, as does the use of frozen autoencoder embeddings although this comes with the benefit of increased throughput during training. 

    \begin{equation}
        \Bbb L' = \Bbb L(O(X, \theta), X_s) + \Bbb L(O(X_c, \theta), X_{sc})
        \label{eq10}
    \end{equation}

    \begin{table}
    \centering
    \begin{tabular}{l c c} 
     \hline
      \textbf{Information Benchmarks} & \textbf{Causal} & \textbf{Combined} \\
      IFEval (strict instance) & 0.2482 & \textbf{0.2506} \\
      LongBench & & \\
      \; \; \scriptsize Code Completion & 13.17\scriptsize±0.31 & \textbf{16.48\scriptsize±0.33} \\ 
      \; \; \scriptsize Few-Shot Learning & 0.80\scriptsize±0.11 & \textbf{1.57\scriptsize±0.12} \\
      \; \; \scriptsize Multi-Document QA & 0.54\scriptsize±0.09 & 0.51\scriptsize±0.09\\
      \; \; \scriptsize Single Document QA & 0.81\scriptsize±0.07 & \textbf{1.07\scriptsize±0.08}\\
      \; \; \scriptsize Summarization & 2.89\scriptsize±0.11 & \textbf{3.41\scriptsize±0.12}\\
      \; \; \scriptsize Information Retrieval & 0.86\scriptsize±0.34 & 0.83\scriptsize±0.37 \\
      SWDE (contains) & 0.0 & \textbf{0.81} \\
     \hline 
     \textbf{Other Benchmarks}  & \textbf{Causal} & \textbf{Combined}\\
      ARC easy & 44.2\scriptsize±1.02 & 43.73\scriptsize±1.02 \\
      WikiText (BPB) & \textbf{1.6752} & 1.6850 \\
      Lambada-OpenAI & \textbf{18.65\scriptsize±0.54} & 17.21\scriptsize±0.53 \\
      HellaSwag & 27.71\scriptsize±0.45 & 27.52\scriptsize±0.45 \\
      \hline
    \end{tabular}
    \caption{Causal versus copy and causal combined objective benchmark accuracy. Both models are frozen encoder-based memory transformers with $d_m=512, n_l=16, n_{ctx}=256, s=4$. All benchmarks are higher is better apart from WikiText.}
    \label{table5}
    \end{table}

    The combined objective training results suggests that memory models are capable of learning to make and decode information-rich encodings and accurately predict next tokens causally, and which task the model performs depends on whether there is a memory delimiter (three special tokens sequentially) or not such that there is no guarantee that the model accurately uses input information during next token prediction. Nevertheless, we find that training a model on the combined objective function leads to notable improvements in input information-related but not general language understanding benchmark performance compared to causal training alone (Table \ref{table5}), indicating that the encoder information is indeed used during causal modeling even though this is not explicitly trained. We note that more benefits would be expected for memory models that use combined objectives that better integrate causal and copy objectives, for instance if the dataset were to contain samples of natural language-based information retrieval.

\section{Discussion}

    \subsection{Information retention in masked mixers and transformers}

\begin{figure}[h]
        \centering
        \includegraphics[width=0.99\textwidth]{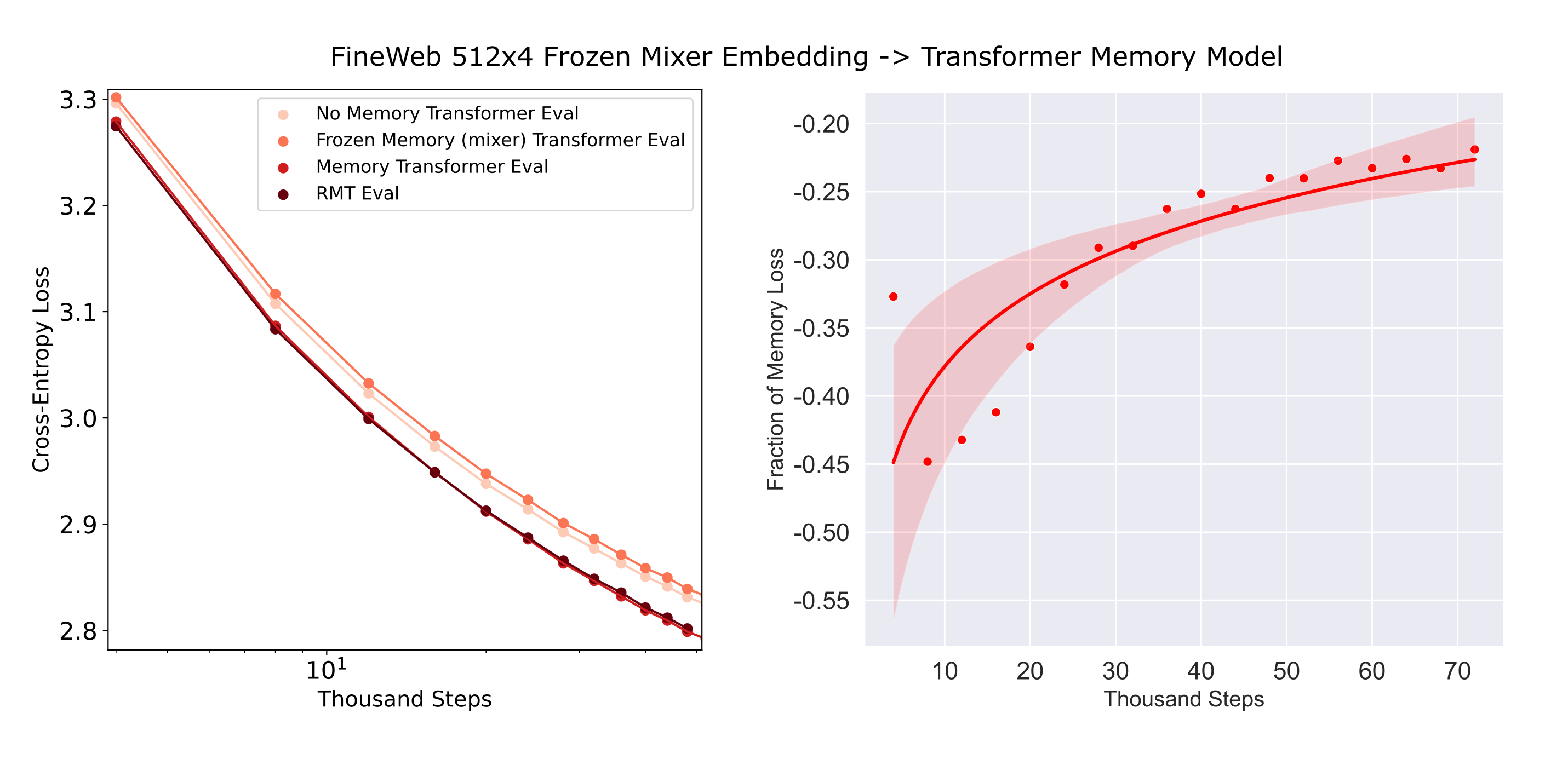}
        \caption{Memory Model Training on FineWeb, Masked Mixer encoder and Transformer decoder ($n_{ctx}=512, s=4, d_m=512, n_l=16$).}
        \label{figs5}
    \end{figure}

    As a general rule, mixers are typically more efficient encoders in terms of information retention and use per compute applied during training, whereas transformers are typically more efficient decoders.  These findings seem to imply that an optimal architecture for modeling using encoders for information retention and decoders for filtering would consist of mixer encoders and transformer decoders, but we find that embeddings from masked mixers are typically poorly compatible with transformer decoders, to the extent that introducing frozen memory model embeddings from masked mixers leads to a decrease in training efficiency relative to non-memory controls (Figure \ref{figs5}). Even when encoders are trainable, we observe a substantial decrease in training efficiency when we pair masked mixer-based encoders with transformer decoders, implying that the manifolds learned by these architectures tend to be substantially different depending on the architecture.

    \subsection{How much information is enough?}

    In this work we have examined information retention using various methods (autoencoder input regeneration, copy, blank copy) and metrics (entropy ratio and token accuracy). A natural question to ask is which method and metric most closely reflects the ability of a model to perform arbitrary input information access. It is clear that token accuracy is a more strict metric than entropy ratio (in that it is always lower for the same model) and that autoencoder input regeneration and blank copy are more strict methods than copy (models with good autoencoder input retention accuracy invariably are capable of accurate copy, but models with accurate copy accuracy do not necessarily exhibit high input information retention or blank copy accuracy). In the context of memory models it may be hypothesized that copy accuracy is most relevant to tasks where less information needs to be accessed, but this remains to be rigorously tested.
    
    \subsection{Implications for Retrieval}

    The finding that language model embeddings are low-information has obvious implications for retrieval. Language model-based retrieval typically occurs by embeddings chunks of text of many target documents, embedding the query, and matching query and target documents using cosine similarity between the query embedding and all targets. Modern retrieval approaches often use multiple embeddings, reranking models, and increasingly approaches such as BM25 to increase the likelihood that the retrieved text matches the target. But as has been noted by \citep{weller2025theoretical} and others, retrieval model accuracy is often much lower than what is theoretically possible for information-rich targets, especially in the case where arbitrary information needs to be retrieved against. Our work provides a clear rationale for these findings: many embedding models used today (regardless of pretraining on causal or masked language modeling objectives) contain only a fraction of input information even for relatively small ($n_{ctx}<256$) chunks, such that arbitrary information in each chunk cannot in general be identified by these models. It remains to be seen whether or not standard retrieval training methods are compatible with information retention.

\bibliographystyle{unsrtnat}
\bibliography{references}  

\beginsupplement
\section{Appendix}

    \subsection{Training Details}\label{methodsection}
    
    We train using models by integrating Pytorch \citep{paszke2019pytorch} defined modules into the Huggingface transformers Trainer \citep{wolf-etal-2020-transformers} utility, adding preprocessing or evaluation processing as required. We train using the torch implementation of AdamW \citep{loshchilov2017decoupled}, typically with a 500 step linear warm up, peak learning rate of $\lambda = 2*10^{-4}$ for Transformers and $\lambda = 5*10^{-4}$ for Mixers (tuned to each model), and with a linear decay for the duration of training, and $n_{ctx}=512$ was used for the context window unless otherwise noted. Causal full-context models, memory models, and autoencoders were typically trained with a global $b=128$ batch size via torch-native distributed data parallel. Models with a context window of $n_{ctx}=1024$ (as was typical for copy model training) were trained with a batch size of $b=64$. For variable-context LLM information retention tests, a constant total number of tokens trained by assigning the batch size to be a function of context length, $b=\floor{32768/n_{ctx}}$, which equates to $b=64$ for $n_{ctx}=512$.

    \subsection{Pretrained LLM memory model training}

    We benchmark for general question answering via Arc-Easy \citep{Clark2018arc}, sentence completion using HellaSwag \citep{zellers2019hellaswagmachinereallyfinish} and Lambada (OpenAI processed) \citep{paperno2016lambadadatasetwordprediction}, and pretraining information retention via MMLU \citep{hendrycks2021measuringmassivemultitasklanguage} using Eleuther-AI's evaluation harness \citep{eval-harness}. Before memory model training, the pretrained Llama 3.2 achieves the benchmark accuracies shown in Table \ref{tables4}, and after training for 32k steps the decoder's benchmark accuracies are shown in Table \ref{tables5}. 

    \begin{table}[!ht]
    \centering
    \renewcommand{\arraystretch}{1.3}
    \begin{tabular}{|l|l|l|l|l|l|l|l|l|}
    \hline
        Tasks & Version & Filter & n-shot & Metric & ~ & Value & ~ & Stderr \\ \hline
        arc\_easy & 1 & none & 0 & acc & ↑ & 0.6633 & ± & 0.0097 \\ 
        ~ & ~ & none & 0 & acc\_norm & ↑ & 0.6170 & ± & 0.0100 \\ 
        hellaswag & 1 & none & 0 & acc & ↑ & 0.4805 & ± & 0.0050 \\ 
        ~ & ~ & none & 0 & acc\_norm & ↑ & 0.6427 & ± & 0.0048 \\ 
        lambada\_openai & 1 & none & 0 & acc & ↑ & 0.6222 & ± & 0.0068 \\ 
        ~ & ~ & none & 0 & perplexity & ↓ & 5.4344 & ± & 0.1288 \\ 
        mmlu & 2 & none & ~ & acc & ↑ & 0.3775 & ± & 0.0040 \\ 
        - humanities & 2 & none & ~ & acc & ↑ & 0.3515 & ± & 0.0069 \\ 
        - other & 2 & none & ~ & acc & ↑ & 0.4303 & ± & 0.0088 \\ 
        - social sciences & 2 & none & ~ & acc & ↑ & 0.4030 & ± & 0.0087 \\ 
        - stem & 2 & none & ~ & acc & ↑ & 0.3394 & ± & 0.0083 \\ \hline
    \end{tabular}
    \caption{Llama 3.2 (1b) benchmarks before copy memory training.}
    \label{tables4}
    \end{table}

    \begin{table}[!ht]
    \centering
    \renewcommand{\arraystretch}{1.3}
    \begin{tabular}{|l|l|l|l|l|l|l|l|l|}
    \hline
        Tasks & Version & Filter & n-shot & Metric & ~ & Value & ~ & Stderr \\ \hline
        arc\_easy & 1 & none & 0 & acc & ↑ & 0.6646 & ± & 0.0097 \\
        ~ & ~ & none & 0 & acc\_norm & ↑ & 0.5976 & ± & 0.0101 \\ 
        hellaswag & 1 & none & 0 & acc & ↑ & 0.4638 & ± & 0.0050 \\ 
        ~ & ~ & none & 0 & acc\_norm & ↑ & 0.6159 & ± & 0.0049 \\ 
        lambada\_openai & 1 & none & 0 & acc & ↑ & 0.5960 & ± & 0.0068 \\ 
        ~ & ~ & none & 0 & perplexity & ↓ & 6.4017 & ± & 0.1668 \\ 
        mmlu & 2 & none & ~ & acc & ↑ & 0.2872 & ± & 0.0038 \\ 
        - humanities & 2 & none & ~ & acc & ↑ & 0.2837 & ± & 0.0066 \\ 
        - other & 2 & none & ~ & acc & ↑ & 0.3238 & ± & 0.0083 \\ 
        - social sciences & 2 & none & ~ & acc & ↑ & 0.2912 & ± & 0.0082 \\ 
        - stem & 2 & none & ~ & acc & ↑ & 0.2525 & ± & 0.0077 \\ \hline
    \end{tabular}
    \caption{Llama 3.2 (1b) Memory Model Decoder benchmark evaluations at 32k copy training steps.}
    \label{tables5}
    \end{table}

\end{document}